\documentclass{article}
\usepackage{ijcai23}

\usepackage{times}
\usepackage{soul}
\usepackage{url}
\usepackage{hyperref}
\usepackage[utf8]{inputenc}
\usepackage[small]{caption}
\usepackage{graphicx}
\usepackage{amsmath}
\usepackage{amsthm}
\usepackage{booktabs}
\usepackage[switch]{lineno}
\usepackage{amssymb} 

\usepackage{subfig}
\usepackage{xcolor}
\usepackage{booktabs}
\usepackage{multirow}

\usepackage[ruled,linesnumbered]{algorithm2e}
\usepackage{algpseudocode}

\title{Addressing Weak Decision Boundaries in Image Classification by Leveraging Web Search and Generative Models}

\author{
Preetam Prabhu Srikar Dammu$^1$
\and
Yunhe Feng$^{1,2}$\And
Chirag Shah$^1$
\affiliations
$^1$University of Washington, Seattle, WA, USA\\
$^2$University of North Texas, Denton, TX, USA
\emails
preetams@uw.edu,
yunhe.feng@unt.edu,
chirags@uw.edu
}

\begin{document}

\maketitle

\begin{abstract}
    Machine learning (ML) technologies are known to be riddled with ethical and operational problems, however, we are witnessing an increasing thrust by businesses to deploy them in sensitive applications. 
    One major issue among many is that ML models do not perform equally well for underrepresented groups. This puts vulnerable populations in an even disadvantaged and unfavorable position. We propose an approach that leverages the power of web search and generative models to alleviate some of the shortcomings of discriminative models. We demonstrate our method on an image classification problem using ImageNet's People Subtree subset, and show that it is effective in enhancing robustness and mitigating bias in certain classes that represent vulnerable populations (e.g., female doctor of color). Our new method is able to (1) identify weak decision boundaries for such classes; (2) construct search queries for Google as well as text for generating images through DALL-E 2 and Stable Diffusion; and (3) show how these newly captured training samples could alleviate population bias issue. While still improving the model's overall performance considerably, we achieve a significant reduction (77.30\%) in the model's gender accuracy disparity. In addition to these improvements, we observed a notable enhancement in the classifier's decision boundary, as it is characterized by fewer weakspots and an increased separation between classes. Although we showcase our method on vulnerable populations in this study, the proposed technique is extendable to a wide range of problems and domains.
    
\end{abstract}

\footnotetext{
    \textbf{Note:} This is a copy of the copyrighted version published in IJCAI 2023 (DOI: \href{https://doi.org/10.24963/ijcai.2023/659}{\color{blue}{10.24963/ijcai.2023/659}}).
}

\section{Introduction}
Computer vision applications have become incorporated into several daily activities in modern societies, and the user base of these applications appears to be growing worldwide as more developing societies are exposed to them. Despite the widespread attention and maturity of the field, this technology and its manifestation in various applications suffers from issues that could have harmful societal impacts. Studies have shown that underrepresentation of certain demographics in datasets imparts bias to machine learning (ML) models \cite{zhao2017men,hendricks2018women,buolamwini2018gender}. This could result in such underrepresented groups becoming more vulnerable, as the negative impacts of these ML services could have far-reaching consequences. Nevertheless, many businesses have rolled out services that rely on flawed technologies in order to expand to untapped markets. 

Businesses often fail to address ML models' performance issues for underrepresented and vulnerable populations because (1) they lack enough resources (primarily, data) required to fairly train their ML models; and/or (2) there may be a concern of how specifically focusing on small groups could negatively affect the performance of large groups, which may bring down the overall accuracy of the models. Technically and economically, it may be prohibitive to have an overall blanketed approach to fix the discrimination problem in an ML model. But, if we could identify specific weakspots in a model and fix them without significantly affecting the rest of the model, we could address this problem of discrimination without sacrificing the overall performance of the model.

In this work, we present a way to leverage generative models and the web to address the challenging task of mitigating bias in services provided to vulnerable populations, which is an essential step towards achieving two of \emph{UN's Sustainable Development Goals (SDGs)}: gender equality (SDG-5) \& reducing inequalities (SDG-10).

To the best of our knowledge, this is the first of its kind attempt to address discrimination against underrepresented (and often vulnerable) classes using a combination of web search and image generation models while also providing a novel framework for enhancing robustness by improving decision boundaries. The rest of the paper is organized as follows. After reviewing some of the related works in Section \ref{sec:related}, we provide an overview of the problem and approach in Section \ref{sec:overview}. The details of our method are presented in Section \ref{sec:methodology}. In Section \ref{sec:datasets}, we describe the datasets used for our experiments, following the experimental details and results in Section \ref{sec:results}. Given that this is a new method for addressing an important problem of bias in ML, we discuss what this means for addressing the needs of vulnerable populations and the UN's SDGs (specifically, SDG-5 and SDG-10) in Section \ref{sec:discussion}. The paper is concluded in Section \ref{sec:conclusion} with some remarks on the current state of this research and future directions.

\section{Related Work}
\label{sec:related}

In this section, we review some of the related concepts and relevant literature required to better understand this work.

\subsection{Bias in Data}
ML models, in general, are built to learn patterns and associations present in the data without questioning their validity and appropriateness. Perhaps, a more concerning finding is that ML models often amplify the bias present in data \cite{wang2019balanced,zhao2017men}. 
To mitigate bias in ML models resulting from imbalanced or inadequate datasets, researchers have proposed several approaches which include balancing datasets in an attempt to address underrepresentation \cite{yang2020towards,Minot2022InterpretableBM,buda2018systematic,d2017conscientious,liu2008exploratory}. 

However, balancing the dataset in terms of the number of samples per class may not be sufficient \cite{Sowik2021AlgorithmicBA,buda2018systematic,wang2019balanced}. For instance, images belonging to the same class might contain information that varies significantly and this may induce biases even when the dataset is balanced. In order to mitigate bias infusing patterns in datasets, it is paramount to identify spurious correlations learned by the ML model and procure training samples that have a neutralizing effect.

\subsection{Robustness in ML Models}
In addition to bias issues, lack of robustness is also a well-known cause for concern when it comes to ML models \cite{carlini2017towards,hendrycks2021natural,taori2020measuring,szegedy2013intriguing}. In machine learning, robustness reflects the model's ability to not being significantly affected by varying conditions. However, a bulk of research has mainly been focused on a specific type of robustness, namely adversarial robustness, which deals with the model's ability to handle adversarial attacks \cite{goodfellow2014explaining}. 
Several scholars have studied the impacts of natural transformations such as changes in lighting conditions  \cite{taori2020measuring,Wang2021RobustTU}. A lesser explored case is the robustness to spurious correlations, which has recently gained more attention \cite{wang2021robustness,wang2021identifying,singla2022salient,plumb2022finding}.

Improving adversarial robustness does not translate to enhanced robustness towards natural transformations \cite{Wang2021RobustTU} or variations arising from distribution shifts \cite{taori2020measuring}. For a model to be reliable, it needs to be robust against varying conditions that arise from the entropy of the real world, and not just from malicious entities.

Disproportionate object to class associations can give rise to spurious correlations, and these patterns compromise the classifier's robustness to distribution shifts \cite{singla2022salient,plumb2022finding}. In \cite{plumb2022finding}, the authors rely on saliency maps and pixel-wise object annotations to identify spurious patterns, and then mitigate these patterns through data augmentation by counterfactual image generation. This method produces classifiers that are more accurate on distributions where the spurious patterns are not helpful and robust to distribution shifts. In contrast, our approach is more generalized as it handles spurious correlations among other shortcomings of the model. 

\subsection{Data Augmentation}
Data augmentation is a widely used technique to address performance issues of ML models. Various approaches to implement data augmentation have been proposed for addressing pitfalls such as class imbalance, overfitting, bias issues, and distribution shifts  \cite{kim2021local,jaipuria2020deflating,yucer2020exploring,hu2019exploring,sharma2020data}. Transformations as simple as rotation or random crop have been proven to improve classifiers \cite{mikolajczyk2018data}. However, in applications where the data distribution is characterized by multiple varying factors, augmentation techniques with higher control over the synthetic augmentation process are required. 

For instance, infinite unique datapoints are bound to exist in an unconstrained real-world environment, which makes capturing long tails of the distribution impractical \cite{jaipuria2020deflating}. To meet such complex requirements, augmenting data through generative techniques such as neural style transfer, GANs, VAEs, and simulation engines have been explored \cite{yucer2020exploring,chen2022real2sim}. However, each of these methods comes with its own set of limitations. Simulation engines serve as a powerful tool if the goal is to diversify scene attributes in robotic tasks \cite{chen2022real2sim}, but are not extendable to use cases beyond the simulated realm.  
Notably, the recent text-to-image generative models \cite{ramesh2021zero,rombach2022high} offer a higher degree of freedom and control in the generation process and have not been explored for data augmentation until now.

\begin{figure*}[t]
  \centering
  \includegraphics[width=\textwidth]{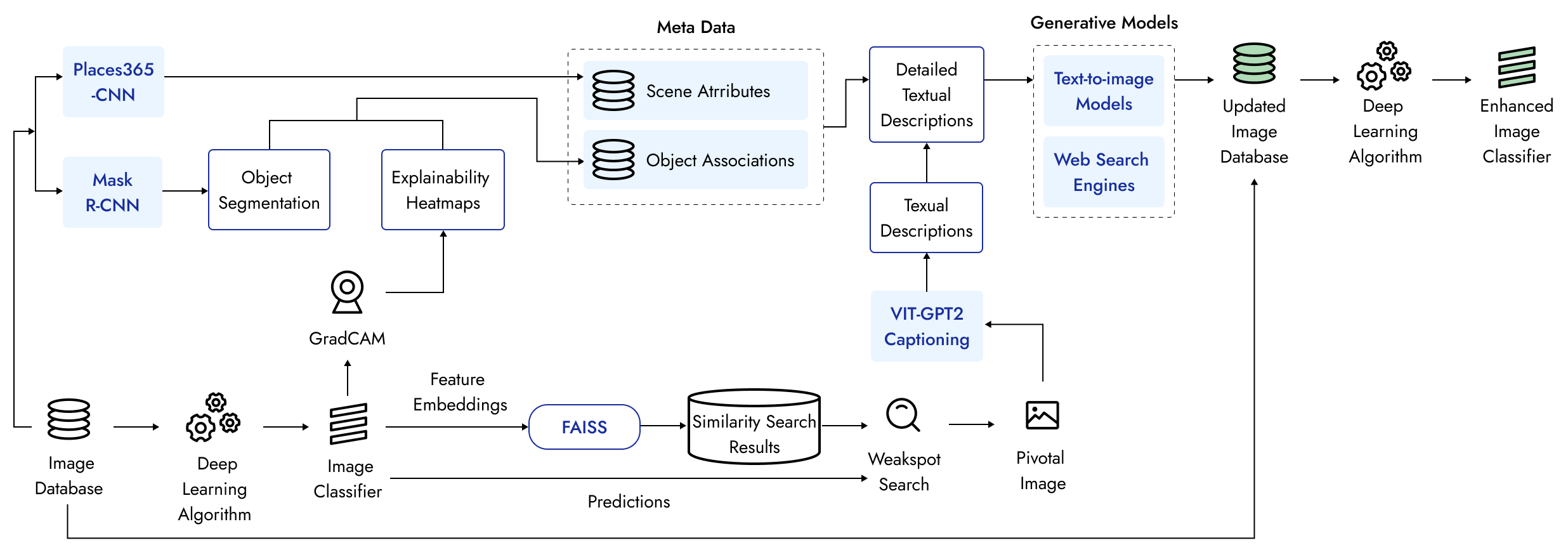}
  \caption{Overview of the proposed approach. Pivotal images representing the weakspots of the classifier are identified and used to generate detailed textual descriptions by leveraging the supporting metadata. New training samples are acquired using these descriptions through generative models which facilitate the enhancement of the classifier (Refer Section \ref{subsec:flowOp}).}
  \label{fig:overview}
\end{figure*}


\section{Overview}
\label{sec:overview}
This section presents the research problem this paper aims to address and an overview of our proposed approach.
\subsection{Research Problem}
Despite various efforts described in the previous section for mitigating bias and robustness issues, we lack a systematic approach that pinpoints where exactly the performance issues for underrepresented classes are coming from and how to address them through data augmentation without disrupting the overall performance of the image classifier.
We break this down into three subproblems: (1) identifying weakspots in the classifier's decision boundary; (2) procuring new datapoints that selectively enhance the decision boundary near the weakspots; and (3) leveraging the augmented data to mitigate the model's bias and enhance its robustness.

\subsection{Approach Overview}
To solve the problem described above, this paper proposes a framework that can automatically detect the weakspots in classifiers and, more importantly, leverage the internet's vastness and the emerging super-realistic text-to-image generative models to mitigate bias and robustness issues. We address all of the subproblems, and thus, our contribution is three-fold.
The overview of the proposed framework is shown in Figure \ref{fig:overview} and its workings are detailed in Section \ref{sec:methodology}.

\section{Methodology}
\label{sec:methodology}
In this section, we present the methodology for identifying the weakspots in the classifier's decision boundary, procuring new training samples that belong to these weak regions with high precision, and remedying the model's robustness and bias issues through strategically captured training data.

\subsection{Identifying Model Weaknesses}

\subsubsection{Identifying Weakspots}\label{subsec:identifyingWeakspots}
To identify weakspots present in the classifier's decision boundary, we need to search the latent space for weak neighborhoods with relatively high perplexity.
However, searching for weakspots in a large latent space is computationally intensive.
Therefore, we adopt a powerful tool that uses GPU acceleration to perform similarity search, Facebook FAISS \cite{johnson2019billion}, thus improving the efficiency of weakspot search. 

A sufficiently large number of data samples that can adequately represent the dataset's distribution are required to identify weakspots.
Feature embeddings are extracted for each image in this representative set, and these feature vectors are fed to the similarity search algorithm.
In our experiments, we use FAISS to perform this step, using the \emph{IndexFlatL2} operation that retrieves top \emph{k} neighbors along with their euclidean distance values for each datapoint.
Subsequently, we perform a grid search on all misclassified instances to check if they lie near a weakspot.
Consider an instance originally belonging to \emph{class 1} erroneously labeled as \emph{class 2}, keeping a maximum neighbor L2 distance \emph{d} as radius, if at least a fixed threshold percentage of neighbors are correctly classified as \emph{class 2}, we detect a weakspot between the two classes in consideration. 
The corresponding misclassified datapoint at the center of the weak region is identified as \emph{pivotal} image.

\subsubsection{Identifying Object Associations and Spurious Correlations}
\label{subsec:spuriousCorr}

\begin{figure}[t]
  \centering
  \includegraphics[width=0.7\linewidth]{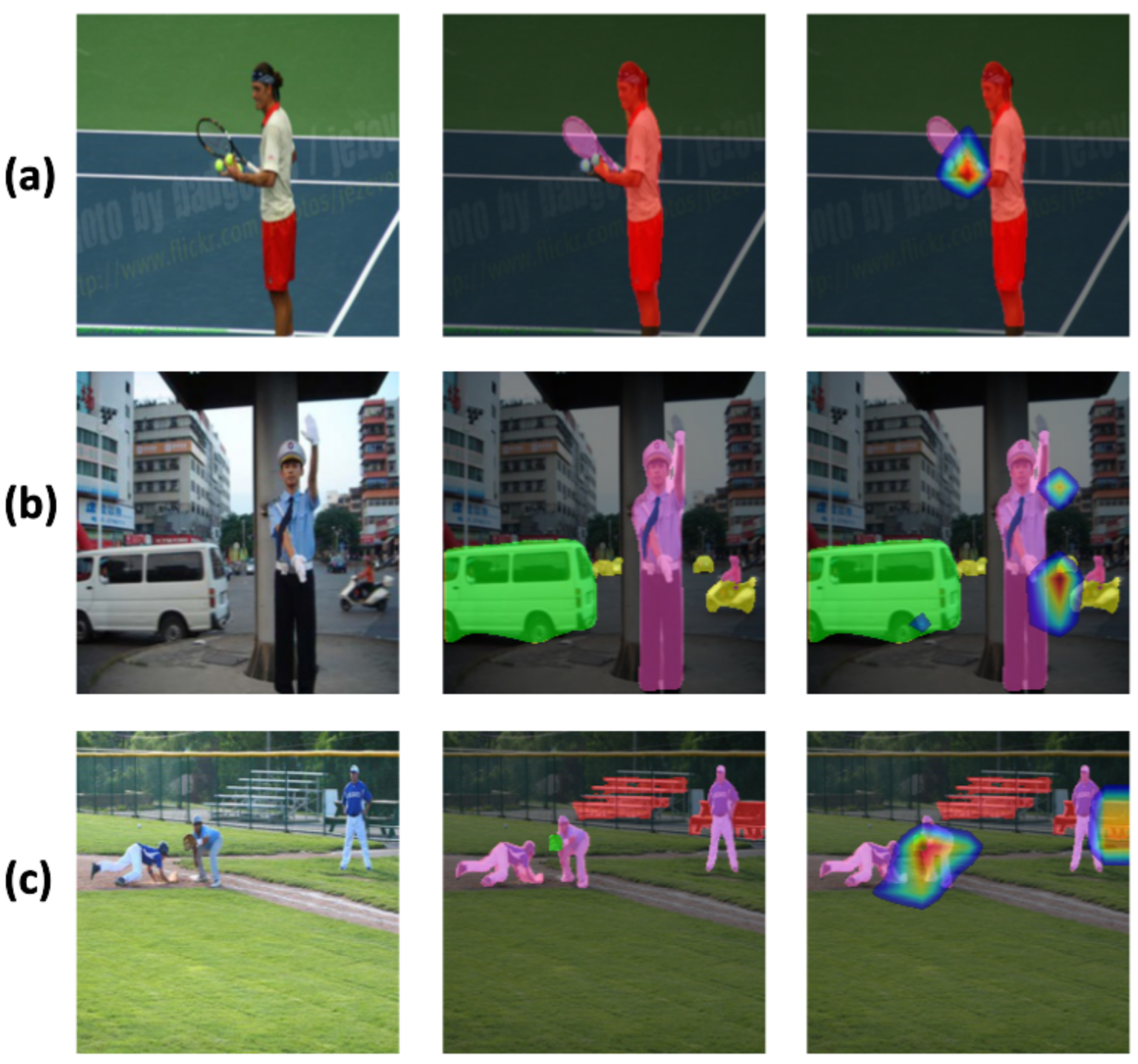}
  \caption{Object associations through heatmaps and segmentation. }
  \label{fig:objectAss}
\end{figure}

We employ a combination of deep learning tasks to identify object associations and spurious correlations learned by the deep learning model.
As understanding the content of an image is an essential first step, we use scene recognition and object detection to obtain necessary image metadata that helps figure out which factor(s) is/are the primary contributors to a classifier's decision.
To achieve this, we rely on explainability heatmaps to detect which objects present in the image appear to trigger the classification.
If the explainability method detects pixels belonging to an object to have higher relevance beyond a certain threshold, an association between that object and the classifier's predicted class is detected for that particular instance.
For example, in Figure \ref{fig:objectAss}, the first column consists of original images, the second column consists of segmented images, and the third consists of heatmaps overlayed on segmented images.
Relevant associations between classes and objects identified in Figure \ref{fig:objectAss} (a) \emph{tennis\_player} -- `tennis racket', `sports ball', `person', Figure \ref{fig:objectAss} (b) \emph{traffic\_cop} -- `person', `car', `motorcycle', `truck', Figure \ref{fig:objectAss} (c) \emph{ballplayer} - `person', `bench', `baseball glove'.

\begin{figure}[t]
  \centering
  \includegraphics[width=1\linewidth]{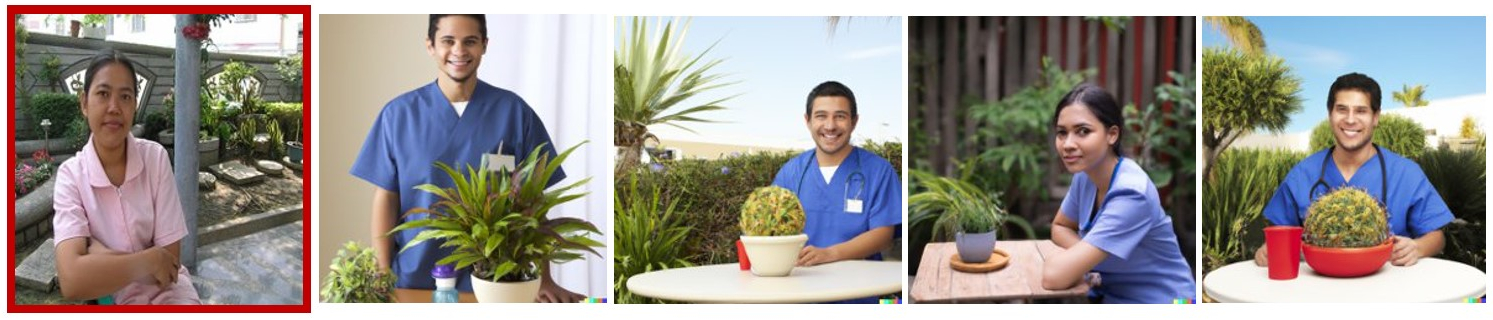}
  \caption{Spurious correlation between \emph{potted plant} and \emph{gardener} class (image shown with red border). Images on the right are images generated using DALL-E to counter this spurious correlation.}
  \label{Fig:spuriousCorr}
\end{figure}

It is non-trivial and highly subjective to decide whether an object association is inappropriate or not.
From the observed associations, the next challenge is to filter out the ones deemed to be spurious.
It requires human judgment, as AI is incapable of making conscious or ethical calls.
Therefore, through manual intervention, we identify questionable associations made by the model.
As the proposed approach identifies scenarios where the model is likely to fail (see Section \ref{subsec:identifyingWeakspots}a), manually checking for the presence of spurious correlations is feasible because the algorithm conveniently shortlists cases which need reviewing. Mitigating search phrases are subsequently added to the set of text prompts to be used for procuring neutralizing images. (see examples in Figure \ref{Fig:spuriousCorr}).

\subsubsection{Addressing Bias Issues}
To mitigate bias, we need to procure samples that neutralize patterns in the data that lead to harmful biases being learned by the ML model. In contrast to standard data balancing, this approach does not merely match the number of samples per class, rather it emphasizes the trends present in the data. This is essential due to the fact that even when the data is perfectly balanced in terms of instances per class, harmful patterns could still be learned \cite{wang2019balanced}. As these patterns lead to weak regions in the decision boundaries, which are detected by our algorithm, effective counterexamples are automatically procured in subsequent steps (see \ref{subsec:procureWSE}b and \ref{subsec:procureGM}c).

\subsubsection{Addressing Robustness Issues}

In this work, we address robustness towards model failures caused by natural triggers such as overfitting on scene information or learning spurious correlations. To address these robustness issues, the identified \emph{pivotal images} representative of weak regions of the classifier are used as `anchoring samples', analogous to `support vectors' in SVMs \cite{chapelle1999support}. Subsequently, images with comparable content to the anchoring samples are either generated using generative models or retrieved using web search engines. Additional datapoints that are sampled from the weakspot's latent space are expected to alleviate the perplexity around it.

\subsection{Fixing Model Weaknesses}
Once we have identified the weakspots in a model, the next step is to find appropriate data to fix them. We do this using web search as well as image generation models.

\subsubsection{Generating Search Phrases for Pivotal Images}

Once weak regions have been identified, we attempt to procure samples from the latent space belonging to these regions.
Pivotal images, as they are located at the centers of these regions with high perplexity, act as a good anchoring point to generate or retrieve similar samples.
As we plan to retrieve samples from web search engines and text-to-image models which take text data as input, an accurate and specific text description of the pivotal image is crucial.
To achieve high-quality descriptions, we use a combination of techniques.

Firstly, we use the Vit-GPT2 \cite{dosovitskiy2020image,radford2019language} image captioning model to generate a caption for the pivotal image.
However, these captions might lack the level of detail to use them for accurately generating new images.
For instance, a common occurrence is that the captions are characterized with pronouns instead of a description of the person in the image.
To remedy this, we replace all pronouns with the class label of the image, as these labels accurately represent the person seen in the image.
Additionally, we use scene information generated by the Places-365 CNN model \cite{zhou2017places} to incorporate scene information into the textual description by only considering the high-level details, such as if the image is taken indoors or outdoors and the venue.
These steps ensure that we obtain a sufficiently detailed and accurate description required to generate or retrieve highly relevant training samples.

\subsubsection{Procuring Images through Web Search Engines}
\label{subsec:procureWSE}
Once detailed descriptions have been generated, web search engines can be used to collect new training images. An advantage of using search engines is that most of the retrieved images can be observed in the real world.
However, in cases where the textual description is of an uncommon instance, the retrieved images may not sufficiently match the search phrase or be irrelevant. For example, the search results of \emph{a person of color female doctor} would be useful, but the results of \emph{a male nurse with a potted plant on a desk} returned by image search engines would not adequately match the description. To address this gap, we generate images with those specific characterizations using generative models.

\subsubsection{Procuring Images through Generative Models}
\label{subsec:procureGM}
In addition to retrieving image search engines, another way to collect desired images is generating them based on given text.
The recently released text-to-image generative models, such as DALL-E 2 \cite{ramesh2021zero} and Stable Diffusion \cite{rombach2022high}, are able to generate high-quality super-realistic images that accurately match the text description of the pivotal image.
Compared to web search engines, text-to-image models allow to sample the weak region more precisely, enabling the generation of highly effective training samples in a more accurate manner.

\subsection{Flow of Operations}
\label{subsec:flowOp}
Here, we present the workflow of our proposed approach in Figure \ref{fig:overview}, and the corresponding algorithm in Algorithm  \ref{alg:generativeAugment}.  
Initially, \emph{Places465-CNN} \cite{zhou2017places} and \emph{Mask R-CNN} \cite{he2017mask} are used to generate scene attributes and segmentation maps respectively for all datapoints, as this metadata is required in subsequent steps. On the image classifier trained on the original training set, we use \emph{GradCAM} \cite{selvaraju2017grad} to generate heatmaps, which are used in conjunction with object segmentation maps to obtain object associations. Next, \emph{FAISS} is used to conduct a similarity search on the feature embeddings to generate the nearest neighbors along with their distances, which are fed to the weakspot search algorithm. The \emph{pivotal images} representing the identified weakspots are then captioned using the \emph{VIT-GPT2} model, and the resulting image descriptions are enhanced with the metadata generated in previous steps to obtain detailed textual descriptions. These descriptions are used for retrieving images using Google image search, as well as for generating images using the text-to-image models. Subsequently, the original dataset is augmented with new datapoints, and this updated dataset is utilized to train the enhanced image classifier.

\begin{algorithm}[tb]
\textbf{Input}: $D_{train}$: train dataset; $D_{test}$: test dataset; $C$: original classifier; $t_{dist}$: L2 distance threshold; $t_{perp}$: perplexity threshold\;

\textbf{Operations}: $\mathbf{proc_{web}()}$: procure images from web; $\mathbf{proc_{txt2img}()}$: procure images from text-to-image model; $\mathbf{finetune()}$: finetune model; $\mathbf{perplexity()}$: compute perplexity; $\mathbf{object\_associations()}$: get object associations; $\mathbf{get\_neighbors()}$: get neighbors within $t_{dist}$; $\mathbf{textual\_description()}$: get textual description; $\mathbf{find\_spurious()}$: find spurious correlations\;

\textbf{Output}: enhanced classifier $C'$\;

$T_{desc} \leftarrow \varnothing$ \tcp*[r]{initialize text descriptions as empty}
$O_{asso} \leftarrow \varnothing$ \tcp*[r]{initialize object associations as empty}

\For{$x_i \epsilon D_{test}$}{
    $neighbors \leftarrow \mathbf{get\_neighbors}(x_i, t_{dist})$ \;
    $perp \leftarrow \mathbf{perplexity}(neighbors)$ \;
    \If(\tcp*[h]{$x_i$ is detected as a pivotal image}){$perp > t_{perp}$}{
        Insert $\mathbf{textual\_description}(x_i)$ into $T_{desc}$\;
        Insert $\mathbf{object\_associations}(x_i)$ into $O_{asso}$\;
    }
}

Insert $\mathbf{textual\_description(find\_spurious}$($O_{asso}$)) into $T_{desc}$\;
$D_{web} \leftarrow \mathbf{proc_{web}}(T_{desc})$ \;
$D_{txt2img} \leftarrow \mathbf{proc_{txt2img}}(T_{desc})$ \;
$D_{updated} \leftarrow D_{train} \cup D_{web} \cup D_{txt2img}$ \;

$C' \leftarrow \mathbf{finetune}(C, D_{updated})$ \tcp*[r]{return enhanced classifier}

\Return $C'$

\caption{Enhancing Classifier}
\label{alg:generativeAugment}

\end{algorithm}

\section{Datasets}
\label{sec:datasets}
\emph{ImageNet People SubTree:} This subset of ImageNet contains 2,832 people categories, however, only 139 of these categories are considered safe and imageable \cite{yang2020towards}. In our experiments, we only considered classes identified as safe and free from annotator bias. From these 139 classes, we selected the ones that are either a profession or an occupation. Categories which share the lowest common hypernym that has a broader yet specific definition of an occupation were merged together. For instance, \emph{captain}, \emph{chief\_of\_staff}, \emph{general}, \emph{major}, \emph{Navy\_SEAL}, \emph{military\_personnel} were all mapped to \emph{military\_officer}, as some of these classes are an abstraction of others. Finally, we ended up with 40 classes that were used to conduct our experiments. Each class had at least 100 images in both training and testing sets.

In \cite{yang2020towards}, the authors prepared annotations for 100 images per each synset, which resulted in 13,900 images. We filtered out categories that were not related to an occupation, and this step resulted in 6,278 images. We used all of the annotated images in the \emph{test} set and the remaining images without annotations in the \emph{train} set, creating a sample size of 6,278 and 64,516 respectively for each partition.

To conduct our experiments, we required further information to help us examine the constituents of the image. To this effect, we used object detection and scene recognition to generate the required additional metadata. 

\begin{figure}[t]
  \centering
  \includegraphics[width=\linewidth]{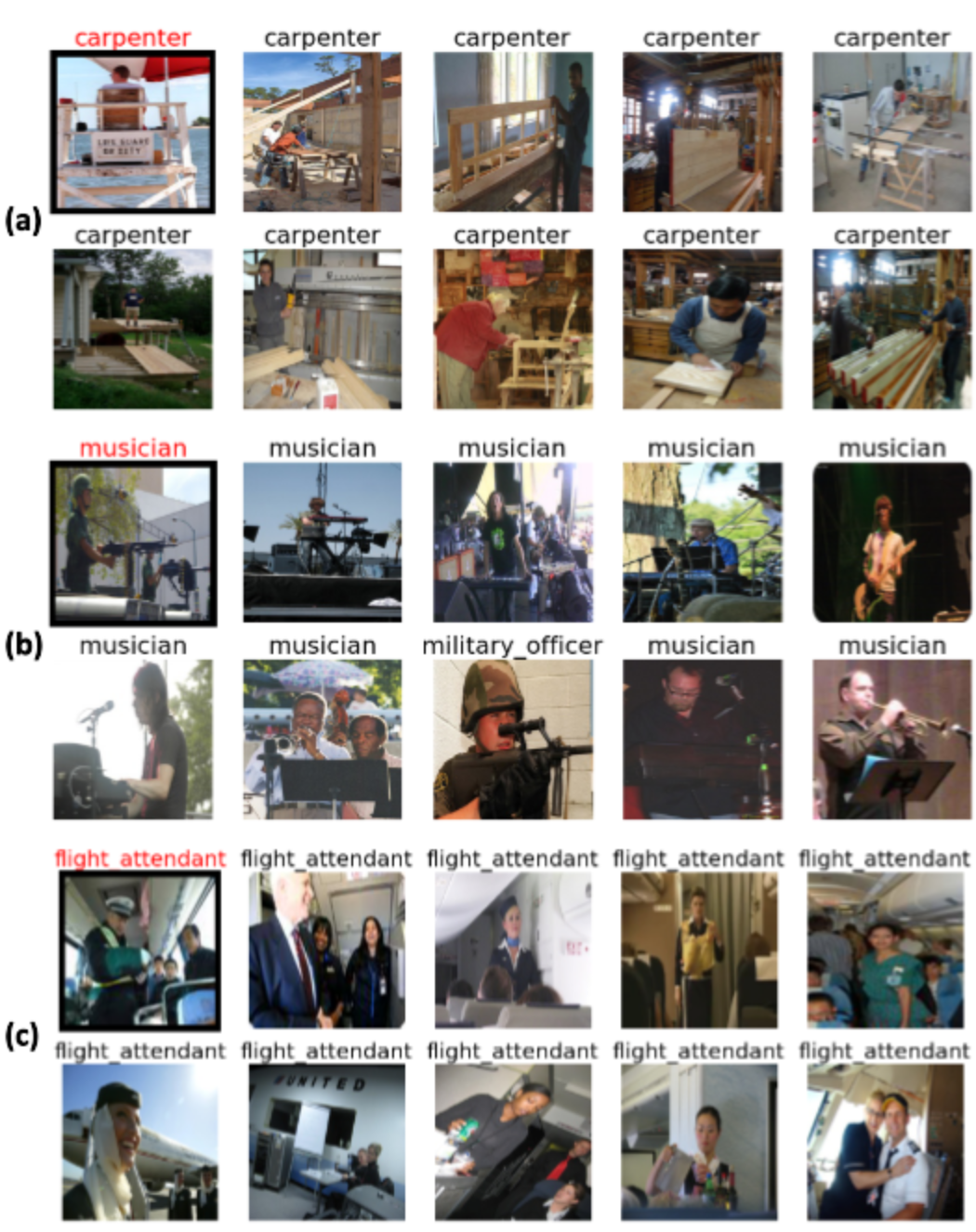}
  \caption{Samples from weak regions between classes (a)  \emph{lifeguard} and \emph{carpenter} (b) \emph{military\_officer} and \emph{musician} (c) \emph{traffic\_cop} and \emph{flight\_attendant}. Erroneous predictions are labeled in red. Images with black borders are \emph{pivotal images}.}
  \label{fig:weakspots}
\end{figure}

\section{Experiments and Results}
\label{sec:results}
To demonstrate the efficacy of the proposed method, we started by building an image classifier on the 40 class subset. Using ResNet50 \cite{he2016deep}, we retrained the last layer on our training set to obtain 80.12\% test accuracy. However, this model demonstrated significant performance disparity. The accuracy for males across all categories was 81.76\%, whereas for females it was 75.68\%, amounting to a 6.08\% gender accuracy disparity. We noticed that the model performed significantly worse for person of color female doctors, as the accuracy for this demographic is just 27.78\% in comparison to 79.38\% for the \emph{doctors} class. After enhancing the classifier with our method, gender accuracy disparity dropped from 6.08\% to 1.38\% amounting to a 77.30\% reduction.

\begin{table}[t]

\includegraphics[width=1\linewidth]{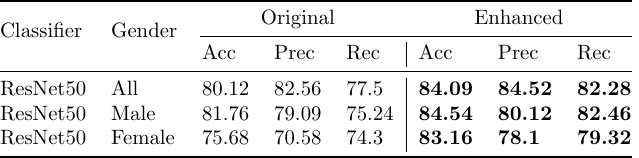}
\caption{Performance of the classifier before and after enhancement.}
\label{perfTable}
\end{table}

\begin{figure}[t]
  \centering
  \includegraphics[width=\linewidth]{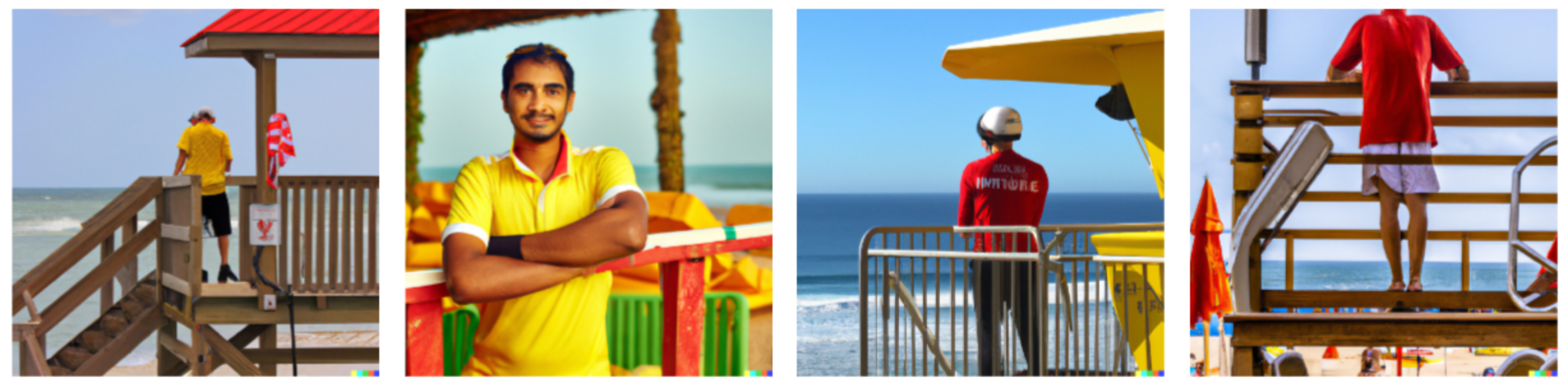}
  \caption{Images generated by DALL-E to mitigate perplexity in weak regions between \emph{lifeguard} and \emph{carpenter} (see Figure \ref{fig:weakspots} (a)).}
  \label{fig:lifeguard}
\end{figure}

\begin{figure}[t]
  \centering
  \includegraphics[width=0.9\linewidth]{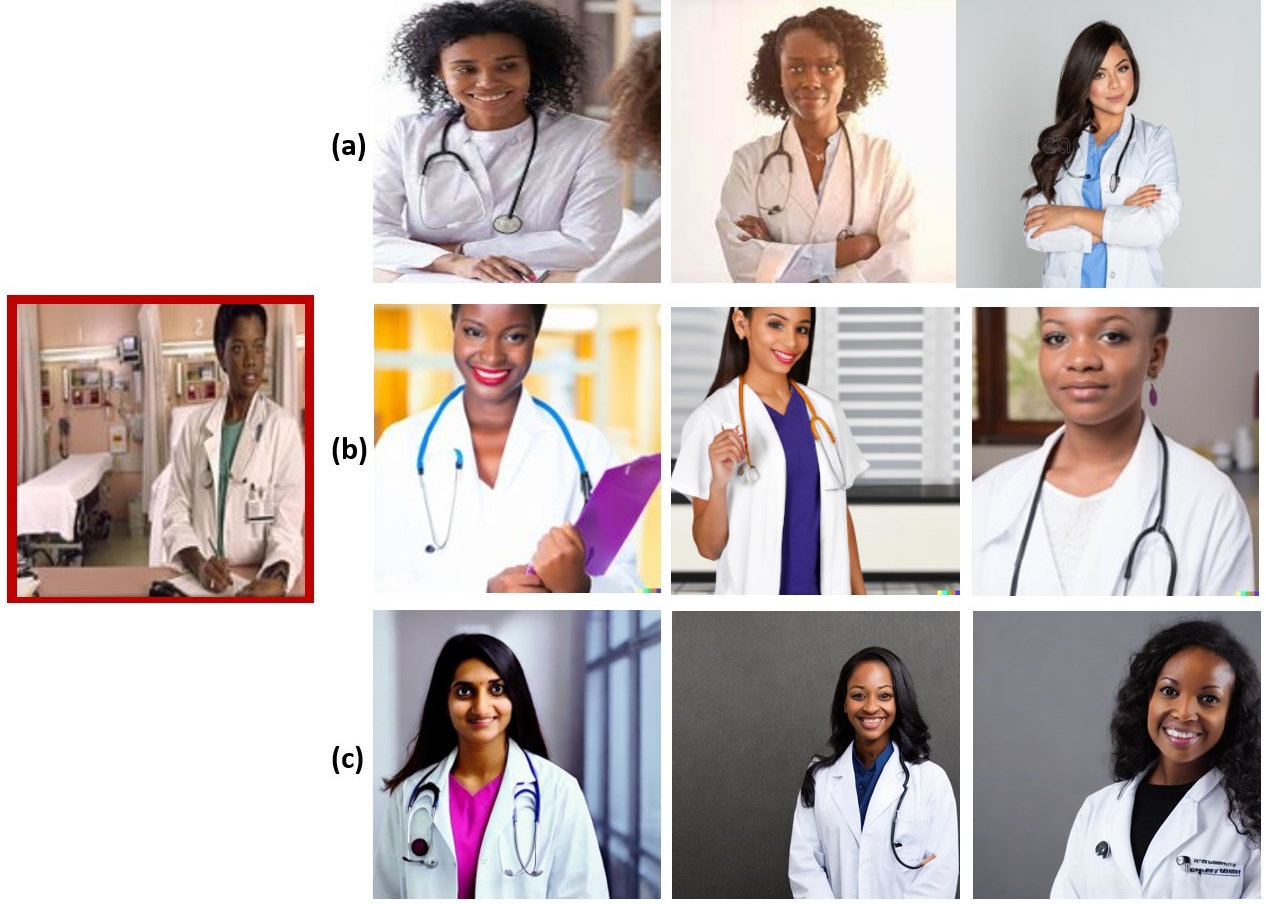}
  \caption{Person of color female doctors have the least accuracy across demographics (27.78\%, overall is 79.38\%). Representative image from ImageNet (left, red borders). Neutralizing images procured through (a) Web Search (b) DALL-E and (c) Stable Diffusion.}
  \label{Fig:femaleDocs}
\end{figure}

\begin{table}[t]
  \centering
  \includegraphics[width=1\linewidth]{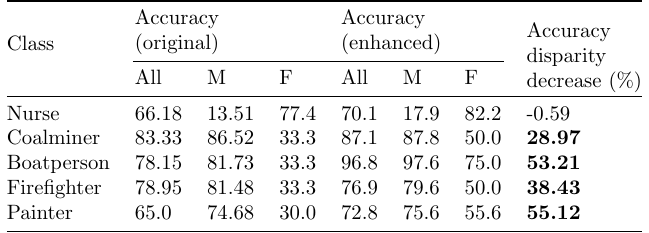}
  \caption{Top five classes with the highest performance disparity for male and female. Significant decrease in accuracy disparity is observed for four of the five classes. Improvements are emboldened.}
  \label{disparityTable}
\end{table}

\begin{table}[t]

\includegraphics[width=1\linewidth]{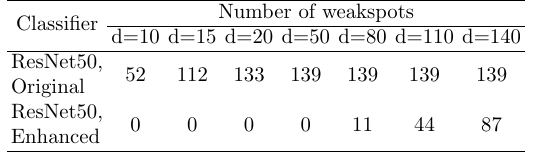}
\caption{Number of weakspots identified with perplexity of 70\%. \emph{d} refers to radius of the weak regions in euclidean distance.}
\label{weakspotTable}

\end{table}

Using our technique, we found weakspots in the decision boundary of the classifiers. Observing the identified weakspots could reveal valuable insights about the model behavior, leading to higher transparency of the automated decision process. Revealing the classifier's weaknesses could also introduce accountability, as the developers of the model would be informed of scenarios where it is likely to fail, and would have to prepare to handle such cases.

For instance, in Figure \ref{fig:weakspots} (a), we notice that a lifeguard is being predicted as a carpenter due to the presence of wooden structures. By sampling near this weakspot between \emph{lifeguard} and \emph{carpenter} classes (see Figure \ref{fig:lifeguard}), we help the model perform better in similar cases. In \ref{fig:weakspots} (b), we see that the model confuses a soldier manning a machine gun with musicians playing an instrument, and the closest three neighboring images share similar scene attributes -- all of them are situated outdoors with open skies. In \ref{fig:weakspots} (c), we observe that a traffic cop is labeled as flight attendant because of the background, as the inside of the bus looks similar to an airplane cabin.

Using the \emph{pivotal images} representing these weak regions, we procured neutralizing images to alleviate the perplexity present around weakspots in the decision boundary. Additionally, we also inspected for any spurious correlations learned by the model by observing the object associations in the identified weakspots. For instance, the presence of \emph{potted\_plant} object in images of nurses fools the model into misclassifying them as \emph{gardener}. To counter this spurious correlation, we generated images of nurses with potted plants situated in front of them using DALL-E. 

Probing performance disparity for the identified weakspots also revealed valuable insights. For instance, taking a closer look at the weakspot between \emph{doctor} and \emph{nurse} classes uncovered the model's bias against the underrepresented demographic of person of color female doctors. The classifier correctly classified doctors 79.38\% of the time, however, it demonstrated a significant drop in accuracy for colored female doctors with an accuracy of 27.78\%. After strategic retraining, the disparity was reduced by 49.37\%.

To neutralize the weakspots identified by the technique, a total of 2,144 neutralizing training samples were procured, increasing the training set size by 3.32\%. Despite being a relatively small-sized addition, the strategically crafted training samples resulted in a considerable improvement in the model's performance. The model's overall accuracy increased by roughly 4\%, but more importantly, the gender accuracy disparity was reduced by 77.30\% (see Table \ref{perfTable}). Top five categories with the highest gender accuracy disparity have been tabulated in Table \ref{disparityTable}, and observe a significant reduction in the disparity for four of the five classes. Notably, this mitigation of bias was achieved without compromising the overall performance of the model.

Sampling weak regions near the decision boundary and retraining with carefully crafted additional datapoints resulted in better-defined class boundaries with fewer weakpoints and better separation. As can be observed in Table \ref{weakspotTable}, the maximum number of weakspots identified in the original model was 139 with \emph{d=50} (L2 distance) as the radius around \emph{pivotal} datapoint in the latent space. After improving the model with the proposed approach, no weakspots were identified at \emph{d=50}, and few were detected at higher \emph{d} values. This indicates a clear increase in inter-class separation and a more robust decision boundary of the classifier.

\section{Discussion}
\label{sec:discussion}
Many ML models suffer from issues stemming from imbalance in datasets \cite{zhao2017men,hendricks2018women,buolamwini2018gender}. Typically, this results in classifiers performing substantially worse for some of its minority classes, even when the overall performance is high \cite{buolamwini2018gender}. The adverse effects of such biases are felt the most by underrepresented and vulnerable demographics, making them susceptible to larger harms of ML-based discrimination. For instance, studies have shown that when it comes to certain job results in image searches, females and people of color are highly underrepresented \cite{lam2018gender}. Even when search engines attempt to address this issue, recent research has shown that the fixes are often on the surface and not foundational \cite{feng2022has}. Such bias in representation leads to cognitive bias, and perpetuates biases in data and algorithms \cite{baeza2018bias}. 

While several studies have attempted to solve the lack of diversity in datasets through data augmentation, previous approaches for generating new samples have limitations  \cite{mikolajczyk2018data,kim2021local,iosifidis2018dealing,jaipuria2020deflating,yucer2020exploring,hu2019exploring,sharma2020data}. A common drawback of most of these approaches is that they are not extendable to other problems, even if they are shown to work well with a specific problem. This is due to the limited degree of freedom and control in the existing approaches. In contrast, our method uses text-to-image generative models that offer a higher degree of variation in multiple aspects such as the background, objects, and person attributes among others.

Any attempt to diversify datasets using traditional approaches such as collecting more data requires a substantial investment in terms of both time and money. In some cases, it might not even be feasible due to operational challenges. There is also a limit to how much diversification can be achieved through additional data collection. However, this should not justify the use of biased datasets to train ML models that may adversely affect vulnerable populations, and the responsibility falls on the ML community to devise solutions that address this challenge. As an alternative, we proposed an approach that could circumvent many of these hindrances by procuring diverse samples instantaneously at low costs. 

Adversarial training typically enhances weak decision boundaries when gradient-based attacks are used to generate the samples, as these techniques compute the least amount of perturbation required to fool the classifier \cite{madry2017towards,goodfellow2014explaining}. However, improvements obtained through this approach are limited to robustness against adversarial attacks and ineffective against natural variations or distribution shifts \cite{Wang2021RobustTU,taori2020measuring}. Moreover, they are not suitable for diversifying the demographics of the training samples. The proposed approach serves an ideal alternative which tackles these shortcomings.

Another major concern about the performance of ML models in deployment is that the distribution of the data it was trained may be different from the real-world data. For example, a medical diagnosis model trained on a predominantly western population may exhibit erroneous behavior when put to practice in other parts of the world. A study on chest X-ray pathology classification model demonstrated this issue, as patients from under-served demographics were underdiagnosed \cite{seyyed2021underdiagnosis,bernhardt2022potential}.

In addition to portability from data distribution to another being an issue, we should also be concerned about the distribution of the same data evolving over time. Therefore, periodical evaluation of machine learning models in deployment is a requirement. As the data changes over time, new weakspots in the decision boundary of the classifier may arise. Consequently, the technique presented in this paper could be used to keep the ML model up to date to reflect natural changes emerging in the distribution of the data.

The proposed approach acts as a step in the right direction to solve the issues discussed above. In the future, we should aim to work towards methods that proactively prevent bias issues, instead of fixing the existing ones in a posthoc fashion.

\section{Conclusion}
\label{sec:conclusion}

In this paper, we presented a method to address three problems: (1) identifying weakspots in image classification; (2) procuring data appropriate for re-learning those weak boundaries; and (3) incorporating such data for training a classifier such that its bias and robustness issues are mitigated.  

The proposed method to enhance discriminative models by leveraging generative models and web search engines instills various desirable characteristics such as robustness, fairness, and  transparency to the original model while still improving the overall performance. 
In addition, the steps involved in executing this approach imparts model understanding and accountability, and as such should be used as a post-development practice before deployment. Finally, this method addresses an often overlooked problem of robustness towards spurious correlations and scene variations. By remedying weakspots through targeted sampling, the decision boundary of the classifier is enhanced with fewer vulnerable points and higher inter-class separation.

While we applied our method on a specific classification problem with a focus on certain vulnerable populations, the method presented in this paper is flexible and can be used to improve classifiers in various applications and domains. As demonstrated, we envision this approach to be extended to several other tasks and promoting better practices in the development of ML models. 

\bibliographystyle{named}
\bibliography{ijcai23}

\end{document}